\documentclass[letterpaper, 10 pt, conference]{ieeeconf}  

\IEEEoverridecommandlockouts                              

\usepackage{amsmath} 
\usepackage{amssymb}  
\usepackage{graphicx}
\usepackage{balance}
\usepackage{wrapfig}
\usepackage{lipsum}
\usepackage{tabularx}
\usepackage{array}
\usepackage{makecell}
\usepackage{booktabs}
\usepackage{subcaption}
\usepackage{xcolor}
\usepackage{hyperref}
\usepackage{cleveref}
\usepackage{amsmath} 
\usepackage{amssymb}  
\usepackage{pifont}
\newcommand{\cmark}{\ding{51}}
\newcommand{\xmark}{\ding{55}}

\usepackage{esvect}
\usepackage{algorithm,algpseudocode}
\usepackage{amssymb} 
\usepackage{multirow}
\usepackage{float}
\usepackage{adjustbox}
\usepackage{booktabs}

\definecolor{nscolor}{rgb}{0.188, 0.478, 0.624}

\algrenewcommand{\algorithmiccomment}[1]{// #1}

\newcommand{\singlequery}{{\mathbf{q}}}
\newcommand{\queryset}{{Q}}
\newcommand{\contact}{{c}}
\newcommand{\repname}{{\text{Object-Centric Contact Flow }}}

\newcommand{\methodfullname}{
SurgFlow:
}

\definecolor{darkgreen}{RGB}{0,150,0}

\title{\LARGE
\methodfullname 3D \repname for Surgical Robot Manipulation
}

\author{Changwei Chen$^1$, Xiao Liang$^1$, Yinuo Yang$^1$, Nicole Shen$^1$, Peihan Zhang$^1$, Sara Wickenhiser$^1$, \\ Zekai Liang$^1$, Soofiyan Atar$^1$, Michael Yip$^1$ 
\thanks{$^{1}$ Department of Electrical and Computer Engineering,
University of California San Diego, La Jolla, CA 92093, USA
{\tt\small \{chc165, x5liang, yiy124, n1shen, pez004, swickenhiser, z9liang, satar, yip\}@ucsd.edu}}
}

\begin{document}

\maketitle
\thispagestyle{empty}
\pagestyle{empty}


\begin{abstract}
Paired video--action demonstrations enable autonomous surgical behavior, but such data is scarce: robots perform roughly 1\% of surgeries, while video-only data is abundant.
Learning 3D object flow offers an embodiment-agnostic way to utilize video data, but flow alone specifies how an object should move, not where and when the tool should engage it, a distinction that is critical in surgery.
We introduce SurgFlow, a framework that learns 3D \emph{\repname} from stereo surgical video without action labels.
For each object point, it predicts a future 3D trajectory and contact scores.
We extract targets via 3D tracking and tool--object proximity, train a flow matching generator to predict them, and use predicted contact to trigger grasp and release while optimizing end effector motion from flow.
On the \textit{da Vinci Research Kit (dVRK)}, SurgFlow succeeds in 37 of 39 stage evaluations across tissue retraction, bimanual reveal, needle pickup, and handover, outperforming baselines trained on equal data with or without action labels.
Zero-shot transfer to a humanoid-based laparoscopic robot achieves 85\% and 70\% average success under similar and novel camera viewpoints, respectively.

\end{abstract}


\section{Introduction}
\label{sec:intro}

Recent data-driven behavioral cloning methods, such as SRT \cite{kim2024surgical} and SRT-H \cite{kim2025srt}, have demonstrated that paired video-action demonstrations provide effective supervision for autonomous behavior in robotic surgeries. 
Yet such data is scarce by construction. Robotic systems perform on the order of 3 million procedures per year \cite{intuitive2026} against an estimated 313 million operations performed worldwide \cite{weiser2015global}: roughly 99\% of surgery involves no robot and therefore produces no robot action labels at all.
Even within robotic surgery, kinematics are rarely recorded or released. 
The largest public action-aligned corpus, assembled from 119 datasets, totals 780 hours \cite{nelson2026open-h}, while the largest prior public single-robot surgical dataset contained around 20 hours~\cite{hansen2026imitatecholec}.
By contrast, individual video-only corpora already approach or exceed this scale (680 hours in GenSurgery \cite{schmidgall2024gsvit} and 938 hours in  LEMON \cite{che2026lemon}), and every one of the hundreds of millions of laparoscopic and open procedures performed each year is a potential source of more video data for robot learning.

Learning from video alone has therefore become an active direction in general and surgical robotics.
One line of work trains text-conditioned video generation models and treats the generated frames as a plan~\cite{he2025surgworld}, but executing that plan still requires an inverse dynamics model trained on paired robot demonstrations, reintroducing the action-data requirement.
Pixel-level prediction is also an inefficient intermediate: it couples motion with texture and lighting, making models computationally demanding and prone to hallucination~\cite{wen2023any}. 
A more compact alternative predicts motion directly as 3D flow~\cite{huang2026pointworld}, which is invariant to viewpoint and directly grounded in scene geometry. 
More recently introduced object-centric representation~\cite{xu2024flow, yuan2024general} tracks flow only on the manipulated object. 
They are embodiment-agnostic: human demonstrations can supply targets that any robot can execute.
However, object flow specifies the object motion, not the tool action that produces it; prior methods thus leave the interaction point implicit and defer it to heuristics or a separate grasp module.
This is tolerable for rigid pick-and-place but not for surgical tasks, where the location and timing of tool--object interaction largely determine success, as shown for needle grasp selection and regrasping~\cite{chiu2021bimanual} and contact-point selection in tool--tissue contact~\cite{liang2025medic}. 
This motivates our research problem: can we learn from video an object-centric representation that captures both object motion and the tool--object contact required to produce it?

\begin{figure}[!t]
    \centering
    \includegraphics[width=\columnwidth]{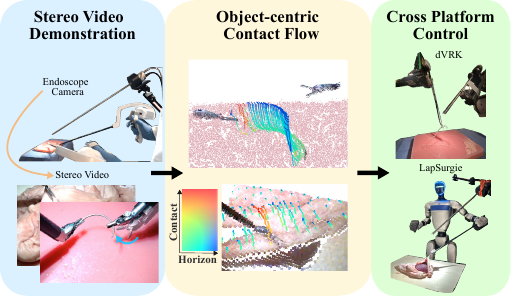}
    \caption{SurgFlow extracts object motion and tool--object contact
    from stereo video demonstrations without robot action labels.
    Contact flow directly guides bimanual surgical manipulation tasks, with experiments on two robot platforms.}
    \label{fig:coverfig}
    \vspace{-15pt}
\end{figure}

\begin{table*}[t]
    \centering
    \caption{Structural comparison of representative methods, as originally proposed. Methods are compared on whether they require action labels, their learned representation and whether it models contact explicitly, how end effector motion and gripper actions are realized, and whether deploying on a new robot platform requires new data.}
    \label{tab:representation_comparison}
    \small
    \setlength{\tabcolsep}{4pt}
    \begin{tabular*}{\textwidth}{@{\extracolsep{\fill}}l c l c l l c@{}}
        \toprule
        & Data & \multicolumn{2}{c}{Learned Representation}
        & \multicolumn{2}{c}{Action Interface}
        & New platform \\
        \cmidrule(lr){2-2}\cmidrule(lr){3-4}\cmidrule(lr){5-6}\cmidrule(l){7-7}
        Method & Action labels & Representation & Explicit contact
        & Motion & Grasping & Requires data \\
        \midrule
        BC/IL~\cite{kim2024surgical,kim2025srt,ze20243d}
        & \textcolor{red}{\cmark} & Implicit & \textcolor{red}{\xmark}
        & Policy & Policy & \textcolor{red}{\cmark} \\
        Video generation~\cite{he2025surgworld, du2023learning}
        & \textcolor{red}{\cmark} & Synthetic frames & \textcolor{red}{\xmark}
        & IDM & IDM & \textcolor{red}{\cmark} \\
        2D flow policy~\cite{wen2023any,xu2024flow}
        & \textcolor{red}{\cmark} & 2D point tracks & \textcolor{red}{\xmark}
        & Policy & Policy & \textcolor{red}{\cmark} \\
        3D flow policy~\cite{lee2026mu_0,lin2026roboflow4d}
        & \textcolor{red}{\cmark} & 3D keypoint flow & \textcolor{red}{\xmark}
        & Policy & Policy & \textcolor{red}{\cmark} \\
        Embodiment flow~\cite{haldar2025point}
        & \textcolor{green}{\xmark} & 3D robot-point flow & \textcolor{red}{\xmark}
        & Optimization & Learned open/close & \textcolor{green}{\xmark} \\
        Object flow~\cite{yuan2024general}
        & \textcolor{green}{\xmark} & 3D object flow & \textcolor{red}{\xmark}
        & Optimization & User-specified & \textcolor{green}{\xmark} \\
        \textbf{SurgFlow}
        & \textcolor{green}{\xmark} & 3D object contact flow & \textcolor{green}{\cmark}
        & Optimization & Contact gating & \textcolor{green}{\xmark} \\
        \bottomrule
    \end{tabular*}
    \vspace{-10pt}
\end{table*}

We introduce SurgFlow, a framework that learns 3D \emph{\repname} from stereo surgical video without any robot action labels.
For each query point on the object, the representation pairs a future 3D motion trajectory with contact scores: the flow specifies how the object moves or deforms, and the contact specifies where and when the tool should engage, hold, or release it as that motion unfolds. 
We extract this representation offline from stereo video by reconstructing and tracking object points in 3D and inferring contact from the spatial proximity between tool and object. 
A flow matching generator trained on these extracted targets then predicts future flow and contact from recent observations. 
Because the prediction is expressed in the object frame rather than in robot actions, it can be executed on any platform through a calibrated controller that turns contact into grasp and release commands and solves for end effector motion that best reproduces the predicted flow. 
This separation between the manipulation behavior
observed in video and the embodiment that executes it is what allows a single learned model to drive different surgical robots. 
Our contributions can be summarized as follows:
\begin{itemize}
    \item We introduce 3D object-centric contact flow, a representation that pairs each object query point's future 3D trajectory with time-varying contact scores to jointly encode object motion and tool--object interaction.
    \item We propose SurgFlow, a framework that combines contact flow extraction, generation, and action optimization to learn from stereo surgical video without robot action labels and execute manipulation across platforms.
    \item We validate SurgFlow across two surgical robot platforms, demonstrating ex-vivo tissue retraction and bimanual needle handover on dVRK and zero-shot transfer of single-arm retraction and bimanual reveal to LapSurgie~\cite{liang2025lapsurgie}, a humanoid-robot-based laparoscopic platform, under different camera viewpoints.
\end{itemize}

\begin{figure*}[!t]
    \centering
    \includegraphics[width=\textwidth]{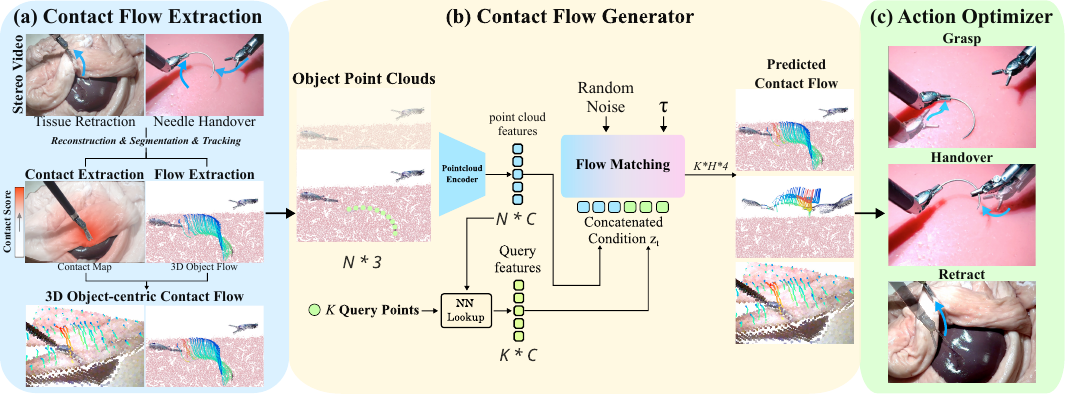}
    \caption{\textbf{Overview of SurgFlow.}
    (a) Tool--object contact and 3D object flow are extracted from stereo video and combined into an object-centric contact flow representation.
    (b) Point cloud features are concatenated with query features to condition a flow matching generator that predicts future contact flow.
    (c) Contact flow supports bimanual manipulation of rigid and deformable objects, from sustained contact during tissue retraction to coordinated grasp and release during needle handover.}
    \label{fig:method_overview}
    \vspace{-10pt}
\end{figure*}

\section{Related Work}
\subsection{Methods for Surgical Automation}
Surgical automation follows two paths. Traditional pipelines pair explicit perception, modeling, and planning: tissue is manipulated with model predictive control over estimated tissue parameters~\cite{shinde2024jiggle}, needle insertion is planned under tissue deformation~\cite{8698220}, and hemostasis is automated with blood detection and suction planning~\cite{richter2021autonomous}. They are precise but rely on hand-designed, task-specific models. Data-driven methods instead learn behavior through reinforcement learning~\cite{yu2024orbit} or, more commonly, imitation learning~\cite{kim2024surgical, haworth2026suturebot, wang2025feedback}. Recent work scales with larger datasets: SRT-H~\cite{kim2025srt}, GR00T-H~\cite{nelson2026open-h}, and SurgWorld~\cite{he2025surgworld}, which still derive supervision from robot action labels. SurgFlow sits between the two: it learns the hardest part of traditional pipelines, how objects move and deform under tool contact, directly from video while keeping execution in explainable models. Unlike learning-based methods that need action labels, it can train directly on stereo surgical videos yet exceeds an action-supervised policy given the same number of demonstrations.

\subsection{Learning Autonomous Behavior from Video}

Table~\ref{tab:representation_comparison} summarizes how methods that learn from video differ in the motion they predict, their need for action labels, and how they decide where and when to grasp.
Among these, video generation methods synthesize task rollouts and then recover actions with an inverse dynamics model~\cite{du2023learning,he2025surgworld}.
Flow policies instead predict more compact 2D or 3D point motion, as in ATM~\cite{wen2023any}, Im2Flow2Act~\cite{xu2024flow}, $\mu_0$~\cite{lee2026mu_0}, and RoboFlow4D~\cite{lin2026roboflow4d}, yet they still rely on policies trained with action labels.
Point Policy~\cite{haldar2025point} removes this requirement by learning from human videos: it predicts 3D robot key points and recovers end effector poses with rigid body constraints, although its gripper states come from hand closure rather than object contact.
Object flow methods also avoid action labels, but they predict only object motion and leave the grasp to heuristics, as in FlowBot3D~\cite{eisner2022flowbot3d}, or to the user, as in General Flow~\cite{yuan2024general}.
In contrast, SurgFlow pairs object flow with explicit contact scores that vary over time and uses them for both motion optimization and coordinated grasp and release, again without any robot action labels.



\section{Method}
\label{sec:method}


\subsection{Problem Formulation}
\label{sec:problem_formulation}
SurgFlow learns to generate \textit{Object-centric Contact Flow} from stereo surgical videos without requiring robot action labels.
This representation jointly encodes object motion and tool-object interaction. 
To make it object-centric, a canonical object frame ${}^w_o\mathbf{T} = (\mathbf{I},\, \texttt{center}({}^wP_0))$ is defined relative to a world frame prior to robot manipulation, with identity orientation and origin at the centroid of the object's initial point cloud ${}^wP_0$. 
Assuming a well calibrated system, all 3D quantities in this work are transformed into and expressed within this canonical object frame. 
The reference frame notations are dropped from now for simplicity.
For a query points $\singlequery\in \mathbb{R}^3$ sampled on the initial object, its contact flow from $t$ to $t+h$ is expressed as
\begin{equation}
    y_{t, h}(\singlequery)
    =
    (\singlequery_{t+h} - \singlequery_{t},
    \contact_{t+h}), 
    \label{eq:contact_coupled_query}
\end{equation}
where the flow is the query point's smoothed displacement and $\contact \in[0,1]$ is a contact score indicating whether the object region represented by $\singlequery$ contacts the tool.
Intuitively, the flow describes how the point should move, while the contact score describes where and when the tool should engage the object.

SurgFlow separates object motion and contact prediction from robot action planning. 
Its first component, a contact-flow generator $f_\theta$ that is based on flow matching \cite{lipman2022flow} (\autoref{sec:flow_generator}), is conditioned on the current and historical point cloud observation $P^o_{t-1}$, $P^o_t$ to predict contact flow for a set of query points $\queryset_t=\{\singlequery^i\}_{i=1}^{K}$ over a horizon of $H$:
\begin{equation}
    \mathcal{Y}_{t}(\queryset_t, H)
    =
    \left\{  y_{t, h}(\singlequery^i)
    \right\}_{\singlequery^i \in \queryset_t,\ h=1,\ldots,H},
    \label{eq:contact_coupled_flow}
\end{equation}
denoted as $\mathcal{Y}_t$ from now on.
Extracting the contact flow representation from raw surgical videos is discussed in \autoref{sec:data_processing}.
The second component is a function $f_c: \mathcal{Y}_t \to \mathbf{a}_t$ that converts generated contact flow to the next robot action \autoref{sec:controller}. 
In SurgFlow, it is implemented as an embodiment specific controller that yields end effector motions and jaw commands based on optimization and gating. 
The separation of the two components allows the same object-centric representation to be used across surgical robotic platforms.

\subsection{Flow and Contact Extraction}
\label{sec:data_processing}

A suite of computer vision algorithms: semantic segmentation \cite{sam2}, stereo disparity estimation \cite{wen2025foundationstereo}, and point tracking \cite{litetracker} are used to extract object-centric contact flow from raw video demonstrations. 
Initial query points $\hat \queryset_0$ are sampled within the object mask via farthest point sampling. 


\textbf{Flow Extraction.} 
Given a demonstration video of length $L$, the initial query points $\hat{\mathcal{Q}}_0$ are tracked across frames using a point tracker, yielding 2D trajectories $\hat{\mathcal{Q}}_{0:L}$.
These are lifted into 3D via inverse projection using known camera intrinsics and estimated stereo disparity. 
To suppress tracking and depth noise, 3D trajectories are smoothed with cubic B-spline fitting. 
The object-centric flow for any query point at time $t$ over horizon $h$ is the displacement between its smoothed 3D position at $t+h$ and its position at $t$.

\textbf{Contact Extraction and Filtering.} 
Contact is estimated from two signals: jaw closure and the proximity distance of object queries to the tool jaws. 
A jaw closure is identified by checking if keypoints on two branches of the jaw are within a certain distance threshold $d_\text{closure}$. 
The effect of proximity is quantified by a spatial evidence score. 
For each query $i$, the evidence is computed as:
\begin{equation}
    e^i_{t}
    =
    \exp\!\left(
    -\frac{\max(\|\singlequery^i_t - \mathbf{p}^\text{jaw}_t\|-d_\text{prox},\,0)}{\lambda}
    \right),
    \label{eq:contact_evidence}
\end{equation}
where $\mathbf{p}^\text{jaw}_t$ is the current jaw mid-point position and $d_\text{prox}$ defines a plateau radius of maximum evidence and $\lambda > 0$ controls how gradually evidence decays beyond it.
Raw spatial evidence is unreliable as a contact value due to observation noise and estimation errors from processing surgical videos. 
Therefore, we additional introduce filtering to compute the final contact values.
Let a closure condition $g_t \in \{0,1\}$ represents jaw closure (1 for closed), and let a per-query visibility condition $v^i_{t} \in \{0,1\}$ be set to $1$ only when query $i$ is visible and its depth is valid and consistent with neighboring measurements. The filtered contact value $c^i_{t}$ that starts initial at zero, updates as
\begin{equation}
    c^i_{t} =
    \begin{cases}
        (1-\alpha)\,c^i_{t-1}+\alpha\,e^i_t, & g_t=1,\ v^i_t=1, \\
        c^i_{t-1},                               & g_t=1,\ v^i_{t}=0, \\
        \gamma\, c^i_{t-1},                      & g_t=0,
    \end{cases}
    \label{eq:contact_filter}
\end{equation}
where $\alpha \in (0,1]$ controls the rate at which new evidence is incorporated and $\gamma \in (0,1)$ governs temporal decay. When the tool is engaged and the query is visible, the score is updated toward the current spatial evidence; during occlusion, it is held at its previous value; and when the tool disengages, it decays smoothly to zero. The resulting contact scores are paired with the extracted object flow for each query to form the training targets in Eq.~\eqref{eq:contact_coupled_flow}.
For bimanual demonstrations, contact is estimated from the jaw geometry and closure states of both tools and represented as a single field over the object, without tool-specific contact labels.


\subsection{Contact Flow Generation}
\label{sec:flow_generator}
The generator conditions on the previous and current object point clouds and the query set. An observation encoder $g_\phi$ maps these to a conditioning token set
\begin{equation}
    \mathbf{z}_t = g_\phi\!\left(P^o_{t-1},\, P^o_t,\, \queryset_t\right),
    \label{eq:obs_encoder}
\end{equation}
and a conditional velocity field $v_\theta$ transports Gaussian noise to the contact flow target. The generator thus defines a conditional distribution rather than a deterministic map,
\begin{equation}
    \mathcal{Y}_t \sim f_\theta\!\left(\,\cdot \mid \mathbf{z}_t\right),
    \label{eq:generator}
\end{equation}
where samples are generated by integrating the ODE defined by the velocity field $v_\theta$ across flow matching timesteps $\tau$.

\textbf{Training and Inference objectives.} 
Each of the four target channels ($3$ flow, $1$ contact) is standardized using its own mean and standard deviation; let $\tilde{\mathcal{Y}}_t$ denote the standardized target. We sample noise $\boldsymbol{\epsilon}\sim\mathcal{N}(\mathbf{0},\mathbf{I})$ and $\tau\sim\mathcal{U}(0,1)$ to form the linear interpolation path
\begin{equation}
    \mathcal{Y}_t^\tau
    =
    (1-\tau)\,\boldsymbol{\epsilon}
    +\tau\,\tilde{\mathcal{Y}}_t,
    \label{eq:flow_matching_path}
\end{equation}
and train $v_\theta$ to regress the constant target velocity $\tilde{\mathcal{Y}}_t-\boldsymbol{\epsilon}$:
\begin{equation}
    \mathcal{L}_{\mathrm{FM}}
    = \mathbb{E}\!\left[
    \big\|
    v_\theta(\mathcal{Y}_t^\tau,\tau,\mathbf{z}_t)
    -(\tilde{\mathcal{Y}}_t-\boldsymbol{\epsilon})
    \big\|_{W}^{2}
    \right],
    \label{eq:flow_matching_loss}
\end{equation}
where $\|\cdot\|_{W}^{2}$ is a weighted squared error, summed over the minibatch, queries, future steps, and channels and normalized by the total weight.

During inference, the conditioning $\mathbf{z}_t$ is computed once. Starting from $\hat{\mathcal{Y}}_t^0\sim\mathcal{N}(\mathbf{0},\mathbf{I})$, we integrate
\begin{equation}
    \frac{d\hat{\mathcal{Y}}_t^\tau}{d\tau}
    =
    v_\theta(\hat{\mathcal{Y}}_t^\tau,\tau,\mathbf{z}_t)
    \label{eq:flow_matching_inference}
\end{equation}
from $\tau=0$ to $\tau=1$ with a fixed-step Euler scheme. 
The target standardization is then inverted to restore displacements to physical units, and contact scores are clipped to $[0,1]$, yielding the prediction $\hat{\mathcal{Y}}_t$ used to generate robot actions.

\textbf{Network Architectures.} 
A shared PointNet++ backbone~\cite{qi2017pointnet++} encodes each cloud into per point features $\mathbf{z}^{\mathrm{pc}}_{t} , \mathbf{z}^{\mathrm{pc}}_{t-1} \in \mathbb{R}^{N \times C}$. 
Per-query features $\mathbf{z}^{q}_{t} \in \mathbb{R}^{K \times C}$ are obtained by interpolating $\mathbf{z}^{\mathrm{pc}}_{t}$ at the query locations $\queryset_t$ from their $k$ nearest neighbors in $P^o_t$. 
The three feature sets are concatenated along the token dimension to form $\mathbf{z}_t$: the two point cloud feature sets supply object geometry and inter-frame motion context, while the query features localize each prediction to its region on the object.

The velocity field is implemented with one token per query. 
For query $i$, the noisy horizon $\mathcal{Y}_{t,i}^\tau\in\mathbb{R}^{H\times4}$ is flattened to $4H$ dimensions, linearly projected to width $C$, and added to $\mathbf{z}^{q}_{t,i}$. 
A DiT-style transformer~\cite{peebles2023scalable} attends across the $K$ query tokens and the conditioning tokens, with $\tau$ injected through adaptive layer normalization.
A shared linear head emits $4H$ values per token, reshaped into an $H\times4$ velocity prediction in standardized flow-contact space. 
Attention exchanges information across queries, while the input projection and feed forward layers mix information across the future horizon within each token.

\textbf{Supervision weighting and augmentation.} Invalid queries, lost tracks, and unavailable labels are excluded from the generator's training loss.
Flow weights are reduced for occluded tracks while retaining valid contact supervision, and scale with the magnitude of ground truth displacement up to a fixed cap so that larger object motions receive greater emphasis. 
During training, rotation augmentation is applied consistently to the point clouds, queries, and displacement targets, and Point-MAE~\cite{pang2023masked} style random masking of local patches simulates partial observations.

\subsection{Contact Flow to Robot Action}
The predicted contact flow $\hat{\mathcal{Y}}_t$ is converted into gripper commands for grasp and release through gating and end effector motions through optimization.
\label{sec:controller}


\textbf{Contact to Gripper Action.} 
When the tool and object are not in contact, their interaction is identified by finding the first time $t$ when a contact value exceeds a grasping threshold $\tau_{\mathrm{on}}$ (i.e. $\exists\, i \text{ s.t. } c^i_t > \tau_{\mathrm{on}}$). 
All grasping valid queries, that satisfy the above condition, are clustered using the fixed-radius clustering method -- two points are in the same cluster if they are both valid and are less than $r_c$ radius apart. 
Grasping clusters are assigned to arms based on proximity. 
For each cluster, the point $q^\text{max}_t$ of highest contact value is selected as the grasp point, and the object type and geometry to determine the grasp orientation.
For tissue, gripper's direction points opposite to the local surface normal at $q^\text{max}_t$. 
During needle handover, we set each gripper's grasp orientation relative to the local needle tangent at $q^\text{max}_t$.
After determining the grasping point and orientation, a gripper approaches $q^\text{max}_t$ with the jaws opened and the grasping orientation fixed, closing the jaws after approached. 
The gripper releases its current engagement when the max contact values of that grasping cluster becomes lower $\tau_{\mathrm{on}}$.

\begin{figure*}[t]
    \centering
    \includegraphics[width=\textwidth]{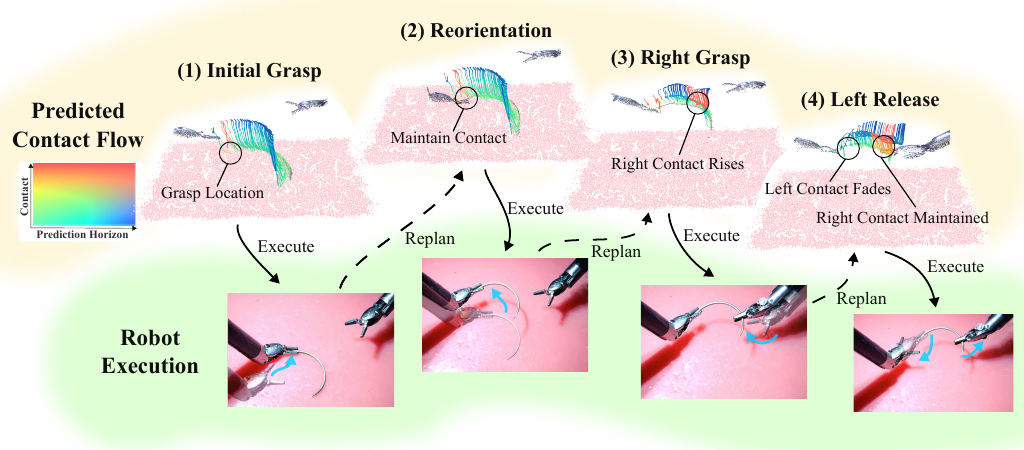}
    \caption{\textbf{Bimanual needle handover with SurgFlow.}
    Predicted contact flow (top) guides execution (bottom) through grasping, reorientation, right grasp, and left release.
    Contact scores remain high at the left grasp during reorientation, then rise at the right before fading on the left to guide the transfer between the two tools.
    Colors encode contact scores and horizon.
    Solid arrows show execution; dashed arrows show replanning from updated observations.}
    \label{fig:needle_handover}
    \vspace{-5pt}
\end{figure*}

\textbf{Flow to End Effector Motion.} When the object is grasped, the controller solves inversely for a chunk of end effector motions that best reproduce the predicted object flow in the vicinity of the grasped region. 
In bimanual execution, each tool independently optimizes its motion using only queries in its assigned contact cluster.
For each future step $t+h$ within the chunk, we solve
\begin{equation}
    \begin{split}
        \Delta R_{t+h}^\star,\, \Delta \mathbf{u}_{t+h}^\star
    =&\\
    \underset{R,\,\mathbf{u}}{\arg\min}
    \sum_{i=1}^{K}
    w^i_{t+h}&
    \left\|
    \mathbf{p}_t + \mathbf{u}
    + R\!\left(\singlequery^i_{t} - \mathbf{p}_t\right)
    - \singlequery^i_{t+h}
    \right\|_2^2,
    \end{split}
    \label{eq:contact_weighted_control}
\end{equation}
where $\Delta R_{t+h}^\star \in SO(3)$ and $\Delta \mathbf{u}_{t+h}^\star \in \mathbb{R}^3$ are the optimal end effector rotation and translation relative to time $t$. The contact-weighted coefficient
\begin{equation}
    w^i_{t+h} = \frac{\hat{c}^i_{t+h}}{\sum_{j=1}^{K} \hat{c}^j_{t+h}}
    \label{eq:contact_weights}
\end{equation}
scales the contribution of each query point proportionally to its predicted contact score, giving queries closer to the grasped region greater influence over the resulting motion. The solution yields a position target $\mathbf{p}_{t+h}^{\mathrm{cmd}}
= \mathbf{p}_t + \Delta\mathbf{u}_{t+h}^{\star}$ and orientation target $R_{t+h}^{\mathrm{cmd}} = \Delta R_{t+h}^\star R_t$, which are executed in sequence before a new prediction is obtained.

\section{Experiments}
\label{sec:exp}

\subsection{Experimental Setup}
\label{sec:exp_setup}

To validate the embodiment-agnostic nature of the proposed method, experiments are conducted across two surgical robot platforms, shown in Fig.~\ref{fig:experimental_setup}: the da Vinci Research Kit (dVRK)~\cite{dvrk} and LapSurgie~\cite{liang2025lapsurgie}, a humanoid-based laparoscopic robot equipped with bipolar forceps and a large needle driver, whose base configuration differs from that of the dVRK.
Both platforms operate in a bimanual configuration with stereo camera perception.
Primary quantitative evaluation is performed on the dVRK, with additional experiments on LapSurgie demonstrating cross-platform transferability.



Evaluation scenarios vary across three axes: contact requirements, material type, and degree of bimanual coordination. The proposed method is evaluated and benchmarked against baselines on four surgical tasks :
\begin{itemize}
\item \textbf{Single Reveal}: A single tissue forcep grasps and lifts intestinal tissue to expose the underlying liver. 
A trial is deemed successful when the liver is visibly exposed.
\item \textbf{Bimanual Reveal}: Two grippers grasp separate regions of a phantom shell and retract in opposing directions to enlarge its opening until the tumor is fully visible.
\item \textbf{Needle Pickup}: A single large needle driver grasps a rigid surgical needle and transports it to a target location. 
Successful pickup requires a stable hold on the needle at the target location.
\item \textbf{Needle Handover}: Continuing from \emph{Needle Pickup}, Arm~1 positions the grasped suture needle for handover while Arm~2 grasps it at a location suitable for subsequent suturing. 
Arm~1 then releases the needle. 
The handover is complete when Arm~2 stably holds the needle at the required location after Arm~1's release.
\end{itemize}

For each task, we collect 100 teleoperated demonstrations on the dVRK, recorded by the stereo camera at a resolution of $640{\times}480$.
DP3 is trained on the same demonstrations with their recorded kinematics, while SurgFlow uses only the stereo videos, with $N{=}1024$ points and $K{=}64$ queries.

\begin{table}[t]
    \centering
    \footnotesize
    \setlength{\tabcolsep}{3pt}
    \renewcommand{\arraystretch}{1.15}
    \caption{
    Task success on the dVRK. 
    Cells report successful trials over attempts from shared randomized initial configurations; 
    handover is evaluated only on trials with successful pickup. 
    Best success rates are shown in bold, including ties; ``---'' denotes an inapplicable setting.}
    \label{tab:task_success}
    \begin{tabularx}{\columnwidth}
        {@{}l*{4}{>{\centering\arraybackslash}X}@{}}
        \toprule
        \textbf{Method}
        & \shortstack{\textbf{Single}\\\textbf{Reveal}}
        & \shortstack{\textbf{Bimanual}\\\textbf{reveal}}
        & \shortstack{\textbf{Needle}\\\textbf{pickup}}
        & \shortstack{\textbf{Needle}\\\textbf{handover}} \\
        \midrule
        DP3
        & 9/10 & 8/10 & 8/10 & 4/8 \\

        Object Flow(w/o contact)
        & 9/10 & 4/10 & 6/10 & --- \\
        
        Tool Flow
        & 9/10 & 7/10 & 8/10 & 0/8 \\
        SurgFlow+IDM
        & \textbf{10/10}
        & \textbf{10/10}
        & \textbf{9/10}
        & 4/9 \\

        \textbf{SurgFlow}
        & \textbf{10/10}
        & \textbf{10/10}
        & \textbf{9/10}
        & \textbf{8/9} \\
   
        \bottomrule
    \end{tabularx}
    \vspace{-10pt}
\end{table}

\subsection{Surgical Task Execution}
\label{sec:exp_main}

\begin{table*}[t]
    \centering
    \small
    \setlength{\tabcolsep}{3pt}
    \renewcommand{\arraystretch}{1.15}
    \caption{Grasp point accuracy on 40 decision states. 
    Errors are reported as median [IQR] in millimeters. 
    $\Delta$ reports the median paired improvement of SurgFlow over each baseline with a bootstrap 95\% confidence interval, and $p_{\mathrm{Holm}}$ denotes the Holm corrected paired Wilcoxon $p$ value.}
    \label{tab:grasp-point}

    \begin{tabular}{@{}lccccccc@{}}
        \toprule
        & \multicolumn{5}{c}{\textbf{Grasp point error (mm), median [IQR]}}
        & \multicolumn{2}{c}{\textbf{Comparison with SurgFlow}} \\
        \cmidrule(lr){2-6}
        \cmidrule(l){7-8}
        \textbf{Method}
        & \textbf{Single Reveal}
        & \textbf{Bimanual Reveal}
        & \textbf{Needle Pickup}
        & \textbf{Needle Handover}
        & \textbf{All}
        & $\boldsymbol{\Delta}$ \textbf{[95\% CI]}
        & $\boldsymbol{p_{\mathrm{Holm}}}$ \\
        \midrule

        Random
        & 34.3 [31.1--41.8]
        & 16.6 [14.7--22.4]
        & 15.3 [10.7--22.2]
        & 5.1 [3.8--9.7]
        & 15.9 [9.7--24.3]
        & 13.4 [8.9--18.7]
        & $3.6{\times}10^{-12}$ \\

        VLM
        & 20.3 [15.9--27.1]
        & 8.2 [6.5--9.7]
        & 15.9 [10.9--18.1]
        & 9.8 [2.3--18.4]
        & 13.7 [7.5--18.4]
        & 9.1 [5.3--14.6]
        & $2.9{\times}10^{-9}$ \\

        Max Flow
        & \textbf{6.7} [4.2--10.4]
        & 10.9 [9.5--14.5]
        & 18.3 [17.4--19.2]
        & 4.6 [1.7--6.2]
        & 10.4 [5.1--16.5]
        & 5.9 [3.1--8.6]
        & $3.5{\times}10^{-7}$ \\

        Tool Flow
        & 6.8 [4.8--13.0]
        & 5.7 [5.2--6.8]
        & 2.4 [1.7--3.2]
        & \textbf{2.2} [1.3--3.1]
        & 4.6 [2.4--6.1]
        & 1.3 [0.5--1.8]
        & $0.007$ \\

        \midrule

        \textbf{SurgFlow}
        & \textbf{6.7} [6.6--7.5]
        & \textbf{3.7} [2.8--3.8]
        & \textbf{1.3} [0.9--1.6]
        & 2.6 [0.7--3.3]
        & \textbf{3.3} [1.4--4.5]
        & ---
        & --- \\

        \bottomrule
    \end{tabular}
    \vspace{-5pt}
\end{table*}


We compare SurgFlow with Object Flow, Tool Flow, SurgFlow+IDM, and DP3 to examine how contact prediction and action generation affect task execution.
\textbf{Object Flow} removes contact prediction and selects grasp locations using the largest predicted flow magnitudes~\cite{eisner2022flowbot3d} while retaining the action optimizer~\cite{yuan2024general}.
\textbf{Tool Flow} follows the rigid motion formulation of ToolFlowNet~\cite{seita2023toolflownet}, which does not directly address the gripper opening and closing required for bimanual needle handover.
For a fair comparison, as the tool approaches its predicted grasp location, a grasp orientation is estimated from object geometry using the same procedure as SurgFlow.
The gripper closes once sufficiently close to the grasp point.
Grasp locations and subsequent manipulation motions remain determined by predicted tool flow.
\textbf{SurgFlow+IDM} uses the predicted object flow and contact but replaces the action optimizer with an inverse dynamics model~\cite{du2023learning,he2025surgworld} trained using robot actions.
\textbf{DP3}~\cite{ze20243d} directly predicts robot actions from 3D observations using paired demonstrations.
\textbf{SurgFlow} combines the same predicted object flow and contact with the action optimizer and does not use robot action labels at any stage of training.


As shown in Table~\ref{tab:task_success}, the methods perform similarly on single arm retraction, but their performance differs as the required contact structure becomes more complex.
Single arm retraction involves one persistent grasp, for which object motion provides a sufficient signal for manipulation.
Bimanual reveal instead requires two spatially distinct contacts to be established and maintained.
Flow magnitude alone does not reliably identify these contacts, leading to incomplete retraction with Object Flow.
Tool Flow also performs well on retraction with geometry-based grasp orientation and proximity-triggered closure, supporting the larger manipulation movements while maintaining an established grasp.

Needle handover further requires both a precise grasp location and an ordered contact transition between the two tools: PSM2 must establish a stable grasp before PSM1 releases the needle.
Although object flow describes how the rigid needle should move, it does not inform timing and location of the grippers' actions.
Tool Flow often completes pickup, but handover exposes two limitations.
First, Tool Flow struggles to recover even the tool’s own orientation accurately from noisy predictions during post-grasp manipulation.
Precise needle manipulation is further complicated by its dependence on the implicit tool–needle grasp relationship, through which tool pose errors can propagate into needle pose errors.
Second, rigid motion decoding does not specify jaw commands, and proximity triggered closure supplies no criterion for release, leaving the required contact transition unspecified.
DP3 and SurgFlow+IDM often complete the initial needle pickup, but their performance degrades during handover, where the receiver grasp and provider release must be precisely coordinated.
SurgFlow remains reliable across the three tasks by coupling the desired object motion with explicit spatial and temporal contact information.

\begin{figure}[t]
    \centering
    \includegraphics[
        width=\columnwidth,
        pagebox=cropbox,
        clip
    ]{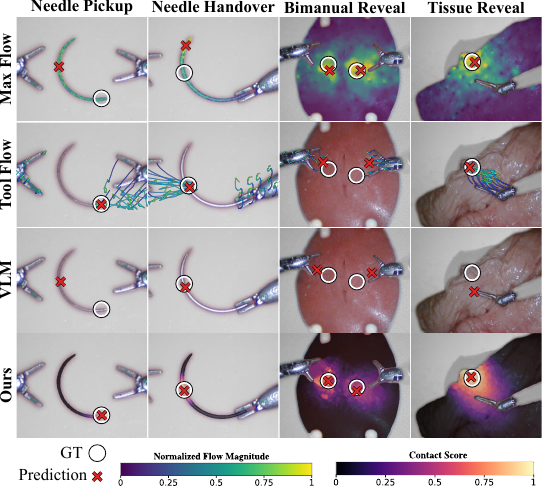}
    \caption{Grasp predictions for needle pickup, handover, bimanual reveal, and tissue retraction (columns) from Max Flow, Tool Flow, VLM, and SurgFlow (rows).
    White circles mark expert annotations; red crosses mark predictions.}
    \label{fig:grasp_location}
    \vspace{-10pt}
\end{figure}

\subsection{Spatial Contact Localization}
\label{contact-localization}

\begin{figure}[t]
    \centering
    \includegraphics[width=\columnwidth]{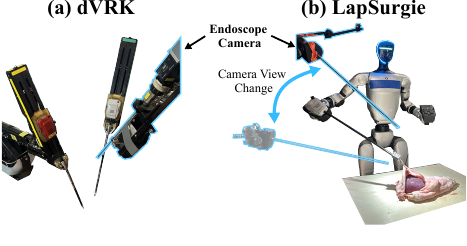}
    \caption{Experimental platforms: (a) dVRK and (b) LapSurgie, with two endoscope viewpoints for cross platform transfer evaluation. Blue outlines indicate the endoscope cameras.}
    \label{fig:experimental_setup}
    \vspace{-15pt}
\end{figure}

To evaluate contact localization independently of the controller, we compare predicted grasp points with expert annotations on 40 held-out examples: ten object configurations for each of four decision states.
Tissue retraction and needle pickup each require one contact, bimanual reveal requires two, and needle handover evaluates the receiver contact.
Error is the Euclidean distance between predicted and annotated contact points for tissue tasks, and the distance along the needle arc for needle tasks.
We report the median and interquartile range per phase, and compare SurgFlow with each baseline using one-sided paired Wilcoxon signed-rank tests with Holm correction, along with the median paired improvement and its bootstrap 95\% confidence interval.


We compare five grasp selection strategies.
\textbf{Random} samples one point from the object point cloud, or two for bimanual reveal, as a reference without task information.
\textbf{VLM} prompts Claude Opus 5 with the endoscopic image and task description, and lifts the predicted pixel to 3D using the same depth map as the expert annotations.
\textbf{Max Flow} follows FlowBot3D~\cite{eisner2022flowbot3d} and grasps where predicted object motion is largest, averaging the top-$k$ queries ranked by flow magnitude at the final prediction step.
\textbf{Tool Flow} rolls out predicted tool motion and selects the nearest object point once a tool comes within a distance threshold, considering only the grasping arm in needle phases.
\textbf{SurgFlow} selects the query with the highest predicted contact score.
For bimanual reveal, Max Flow and SurgFlow cluster top-ranked queries into two groups and keep the strongest in each.

As shown in Table~\ref{tab:grasp-point} and Fig.~\ref{fig:grasp_location}, SurgFlow achieves the lowest overall contact localization error and remains consistent across the four decision states. 
The Random baseline varies substantially across phases because the measured error depends on both object size and the position of the annotated target, making it a phase specific reference rather than a direct comparison across tasks.
VLM produces large and variable errors, indicating that semantic reasoning alone lacks the geometric precision for contact selection.
Max Flow performs well for single arm retraction, where the region of largest tissue motion is also a suitable grasp region. 
Its accuracy decreases for bimanual reveal because flow magnitude does not precisely identify two distinct contacts. 
It performs particularly poorly for needle pickup, where rotational motion can produce large displacement at the needle endpoint even when the desired grasp region moves very little. 
Tool Flow is the closest baseline and slightly outperforms SurgFlow during handover.
However, its contact estimate is inferred indirectly from the predicted tool trajectory, making it sensitive to errors accumulated during rollout. 
SurgFlow instead predicts contact directly on the object, achieving the best results for bimanual reveal and needle pickup while matching Max Flow for single arm retraction.
Across all 40 decision states, the paired improvement over every baseline remains significant after Holm correction, and each bootstrap confidence interval excludes zero.

\subsection{Transfer and Generalization}
In robot-assisted surgery, robot configurations vary across setups, and the camera viewpoint may change during a procedure.
We therefore evaluate whether SurgFlow's object-centric representation supports zero-shot transfer to a different robot and remains robust to viewpoint changes.
On LapSurgie, we test single arm ex vivo tissue retraction and bimanual reveal under two calibrated camera views: one similar to the training viewpoint, and one more extreme with greater object self-occlusion (\autoref{fig:experimental_setup}).
\textbf{SurgFlow} uses the pretrained generator and action optimizer without fine-tuning.
\textbf{DP3} transfers the dVRK trained policy using LapSurgie's end effector states and perception.
\textbf{SurgFlow w/o augmentation} is trained without occlusion augmentation and evaluated under the extreme view.

Table~\ref{tab:transfer_generalization} reports task success on LapSurgie for each method and camera condition.
SurgFlow transfers to LapSurgie without fine-tuning and achieves higher success on single-arm retraction than on bimanual reveal.
Most failures occur during reaching because LapSurgie's controller cannot position the tools accurately enough to establish the planned contacts.
The DP3 policy trained on dVRK completes neither task after direct transfer to LapSurgie.
Under the extreme view, SurgFlow's success rate drops more on bimanual reveal than on single arm retraction, while removing occlusion augmentation leads to a much larger decline on both tasks.
The object-centric frame gives observations from different camera poses a common reference, and the ablation suggests that augmentation helps the model handle self-occlusion by exposing it to partial object geometry during training.


\begin{table}[t]
\centering
\footnotesize
\setlength{\tabcolsep}{3pt}
\renewcommand{\arraystretch}{1.15}
\caption{Task success on LapSurgie, reported as successful trials
out of ten attempts. Ex-vivo retraction uses one arm; reveal uses
both arms. The augmentation ablation uses SurgFlow trained without
occlusion augmentation.}
\label{tab:transfer_generalization}
\begin{tabular}{@{}lcccc@{}}
\toprule
Task
& \shortstack{SurgFlow\\Similar view}
& \shortstack{SurgFlow\\Extreme view}
& \shortstack{w/o aug.\\Extreme view} 
& DP3\\
\midrule
Bimanual reveal    & 7/10  & 5/10 & 1/10 & 0/10 \\
Ex-vivo retraction & 10/10 & 9/10 & 3/10 & 0/10 \\
\bottomrule
\end{tabular}
\vspace{-10pt}
\end{table}

\section{Limitations and Conclusion}
In this work, we estimate contact from jaw closure and tool--object proximity, which may not always reflect actual engagement.
More accurate surface reconstruction and tool pose estimation could improve these contact labels.
Our representation also does not fully specify the relative orientation between the tool and object, so grasp orientation still relies on geometric heuristics.
Finally, calibration errors and limited controller accuracy can prevent the tools from reproducing the desired motion, especially when transferring to a new platform with less precise control.
We hope to address these limitations in future work to support more precise and varied surgical manipulation.

In summary, we demonstrated surgical manipulation from stereo video without robot action labels by jointly representing object motion and tool--object contact.
SurgFlow guides motion and grasp--release coordination on dVRK and transfers retraction tasks zero-shot to LapSurgie, supporting broader use of surgical video for robot learning.

%

\bibliographystyle{IEEEtran}
\bibliography{root}

\end{document}